\documentclass[preprint,12pt]{elsarticle}
\usepackage{textcomp}
\usepackage{stfloats}
\usepackage{verbatim}
\usepackage{cite}
\usepackage{color}
\usepackage{xcolor}
\usepackage{graphicx}
\usepackage{amsmath}
\usepackage{amssymb}
\usepackage{booktabs}
\usepackage{verbatim}
\usepackage{bbold}
\usepackage{multirow}
\usepackage{makecell}
\usepackage{pifont}
\usepackage{algorithm}
\usepackage{algorithmic}
\usepackage{setspace}
\usepackage[caption=false,font=normalsize,labelfont=sf,textfont=sf]{subfig}
\usepackage{caption}

\makeatletter
\def\hlinew#1{%
  \noalign{\ifnum0=`}\fi\hrule \@height #1 \futurelet
   \reserved@a\@xhline}
\makeatother

\usepackage[pagebackref,breaklinks,colorlinks]{hyperref}
\usepackage[capitalize]{cleveref}
\crefname{section}{Sec.}{Secs.}
\Crefname{section}{Section}{Sections}
\Crefname{table}{Table}{Tables}
\crefname{table}{Tab.}{Tabs.}

\usepackage{newpxtext}
\newcommand{\ie}{\emph{i.e.}{}}

\newcommand{\eg}{\emph{e.g.}{}}
\newcommand{\etc}{etc\@ifnextchar.{}{.\@}}

\usepackage[numbers]{natbib}

\usepackage{amssymb}

\journal{arXiv}

\begin{document}

\begin{frontmatter}



\title{Learning with Noisy Labels Using Collaborative Sample Selection and Contrastive
Semi-Supervised Learning}

\author[label1]{Qing Miao}
\author[label2]{Xiaohe Wu}
\author[label1]{Chao Xu}
\author[label3]{Yanli Ji}
\author[label2]{Wangmeng Zuo}
\author[label4]{Yiwen Guo}
\author[label1]{Zhaopeng Meng}

\affiliation[label1]
{organization={College of Intelligence and Computing, Tianjin University},
            city={Tianjin},
            postcode={300072}, 
            country={China}}

\affiliation[label2]
{organization={School of Computer Science and Technology, Harbin Institute of Technology},
            city={Harbin},
            postcode={150006}, 
            country={China}}
\affiliation[label3]
{organization={Center for Future Media, School of Computer Science and Engineering, UESTC},
            city={Chengdu},
            postcode={611731}, 
            country={China}}
\affiliation[label4]
{organization={independent researcher},
            country={China}}
\begin{abstract}
Learning with noisy labels (LNL) has been extensively studied, with existing approaches typically following a framework that alternates between clean sample selection and semi-supervised learning (SSL). However, this approach has a limitation: the clean set selected by the Deep Neural Network (DNN) classifier, trained through self-training, inevitably contains noisy samples. This mixture of clean and noisy samples leads to misguidance in DNN training during SSL, resulting in impaired generalization performance due to confirmation bias caused by error accumulation in sample selection.
To address this issue, we propose a method called Collaborative Sample Selection (CSS), which leverages the large-scale pre-trained model CLIP. CSS aims to remove the mixed noisy samples from the identified clean set. We achieve this by training a 2-Dimensional Gaussian Mixture Model (2D-GMM) that combines the probabilities from CLIP with the predictions from the DNN classifier.
To further enhance the adaptation of CLIP to LNL, we introduce a co-training mechanism with a contrastive loss in semi-supervised learning. This allows us to jointly train the prompt of CLIP and the DNN classifier, resulting in improved feature representation, boosted classification performance of DNNs, and reciprocal benefits to our Collaborative Sample Selection.
By incorporating auxiliary information from CLIP and utilizing prompt fine-tuning, we effectively eliminate noisy samples from the clean set and mitigate confirmation bias during training. 
Experimental results on multiple benchmark datasets demonstrate the effectiveness of our proposed method in comparison with the state-of-the-art approaches.

\end{abstract}



\begin{keyword}
 Learning with noisy label, collaborative sample selection, contrastive loss, semi-supervised learning.



\end{keyword}

\end{frontmatter}


\section{Introduction}
\label{sec:intro}
Deep neural networks (DNNs) have remarkable advancements in computer vision tasks~\citep{joseph2021towards,krizhevsky2017imagenet,yang2021objects,ge2020cascaded,wang2021contrastive}.
The superior performance relies on large-scale datasets that are meticulously annotated with high-quality labels.
Nevertheless, the tremendous quantities of correct annotations are challenging and time-consuming to collect.
Therefore, crowd-sourcing and web searching are used for data annotation, which inevitably involve noisy labels.
Due to the memorization effect of DNNs, noisy labels impair their performance seriously~\citep{arpit2017closer,liu2020early}.
Consequently, learning with noisy labels is widely explored and researched in literature~\citep{jiang2018mentornet,shu2019meta,song2019selfie,yi2019probabilistic,li2020dividemix,propmix,zheng2021meta,zhang2019metacleaner,chen2022compressing}.

\begin{figure}[t]
  \centering
  \hspace{4.5mm} 
  \vspace{-1.5mm}
  \small{CIFAR-100 with 90\% symmetric noise}
\includegraphics[width=0.75\linewidth]{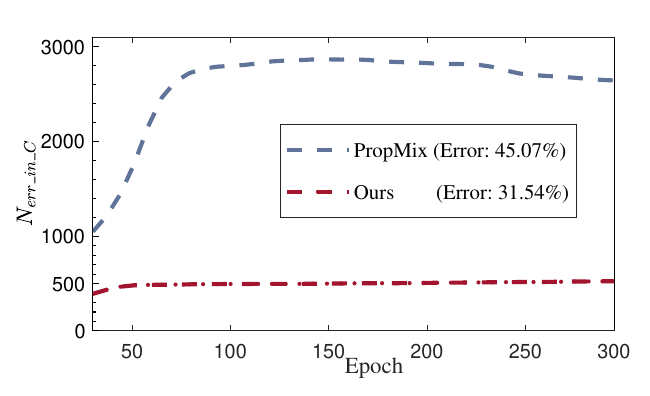}
   \caption{Change of the number of mislabeled samples mixed in the clean set, \ie, $N_{err\_in\_C}$, over the training $Epoch$. $Error$ refers to the test error of the last epoch model. PropMix\citep{propmix} divides a large number of mislabeled samples into the clean set from the start. This data selection bias gives rise to model confirmation bias, whereby the model progressively adapts to the mislabeled samples with DNN training, consequently leading to degraded generalization performance on the test set (45.07\%). However, our method effectively constrains the error accumulation in sample selection and reduces the model confirmation bias with a test error of 31.54\%.}
   \label{fig:1}
\end{figure}

To combat label noise, existing solutions primarily emphasize on alternating processes, including a sample selection approach and a semi-supervised learning (SSL) method~\citep{li2020dividemix,propmix,karim2022unicon,wei2022self}.
However, some noisy samples are inevitably distinguished into the clean set in sample selection using the self-generated information of the classifier trained by SSL, \eg, loss and predictions.
Then, these noisy samples in the clean set in turn severely impair the performance of the classifier via SSL.
The confirmation bias is induced by the alternating process between sample selection and semi-supervised learning.
Moreover, we observe that during the iterative learning, the noisy samples mixed in the divided clean set are accumulated, as shown in the blue curve of Figure~\ref{fig:1}.
Along with semi-supervised learning, lots of noisy samples would be mixed in the clean set and cannot be accurately discriminated.
Finally, it severely deteriorates the generalization performance of the classifier.

To address the confirmation bias, we aim to introduce auxiliary information to assist in the clean sample selection.
Recently, the large-scale pre-trained models trained on large image-text pairs with language supervision have shown great potential in feature representations~\citep{radford2021learning,li2019visualbert,jia2021scaling,kim2021vilt}.
It provides powerful representation capacity and shows potential in diverse downstream tasks via zero-shot or few-shot learning~\citep{patashnik2021styleclip,li2021align,rao2022denseclip,ding2022open}.
Therefore, in this paper, we propose to leverage the large-scale pre-trained model, \ie, CLIP, to assist in the traditional DNNs classifier in clean sample selection.
With the auxiliary information from CLIP, the clean samples can be co-rectified and selected accurately, especially when the dataset is corrupted with severe label noise. 
Figure~\ref{fig:1} illustrates the number of noisy samples mixed in the selected clean set at different epochs on CIFAR-100 with 90\% symmetric noise, and provides the test error of the last epoch model.
It can be noted that, compared with PropMix~\citep{propmix}, the number of noisy samples mixed in the clean set divided by our approach is effectively constrained, and the trend of noise accumulation is significantly reduced.
PropMix achieves a test error rate of 45.07\%, whereas our approach maintains an error rate around 31.54\%.
In this paper, we propose a novel Collaborative Sample Selection (CSS) method in cooperation with the pre-trained model CLIP, which helps to filter out the noisy samples mixed in the selected clean set, and ensures the quality of the clean set for the subsequent semi-supervised learning.
Moreover, to improve the performance and adaptability of CLIP and classifier, we employ a co-training mechanism to fine-tune the prompt of CLIP and train the DNNs classifier jointly in the SSL stage.
In the implementation, we also include the unsupervised contrastive loss and semi-supervised loss to improve the robustness of the feature representation~\citep{he2020momentum,chen2020improved,chen2020simple,chen2020big}.
Based on extensive experimentation conducted on CIFAR-10/CIFAR-100 and real-world datasets, our method has consistently shown superior performance when compared to state-of-the-art approaches. 
We take pride in highlighting the following significant contributions of our work:
\begin{itemize}
    \item We propose a Collaborative Sample Selection method, \ie, CSS, in cooperation with the pre-trained model CLIP, to ensure the quality of the divided clean set. To reduce confirmation bias caused by error accumulation in the selected clean set, we combine CLIP probabilities and DNN classifier predictions using a 2D-GMM model. This ensures accurate identification of clean samples.    
    %
    \item We adopt a co-training mechanism in the semi-supervised learning (SSL) stage to fine-tune the prompt of CLIP and train the DNNs classifier jointly. It not only improves the adaptability and performance of CLIP and DNNs respectively, but also boosts our Collaborative Sample Selection in turn.
    \item We are currently conducting extensive experiments on synthetic and real-world noisy datasets and the preliminary results indicate that our approach outperforms the state-of-the-art methods.
\end{itemize}

\begin{figure*}[t]
\centering
   \includegraphics[width=0.98\textwidth]{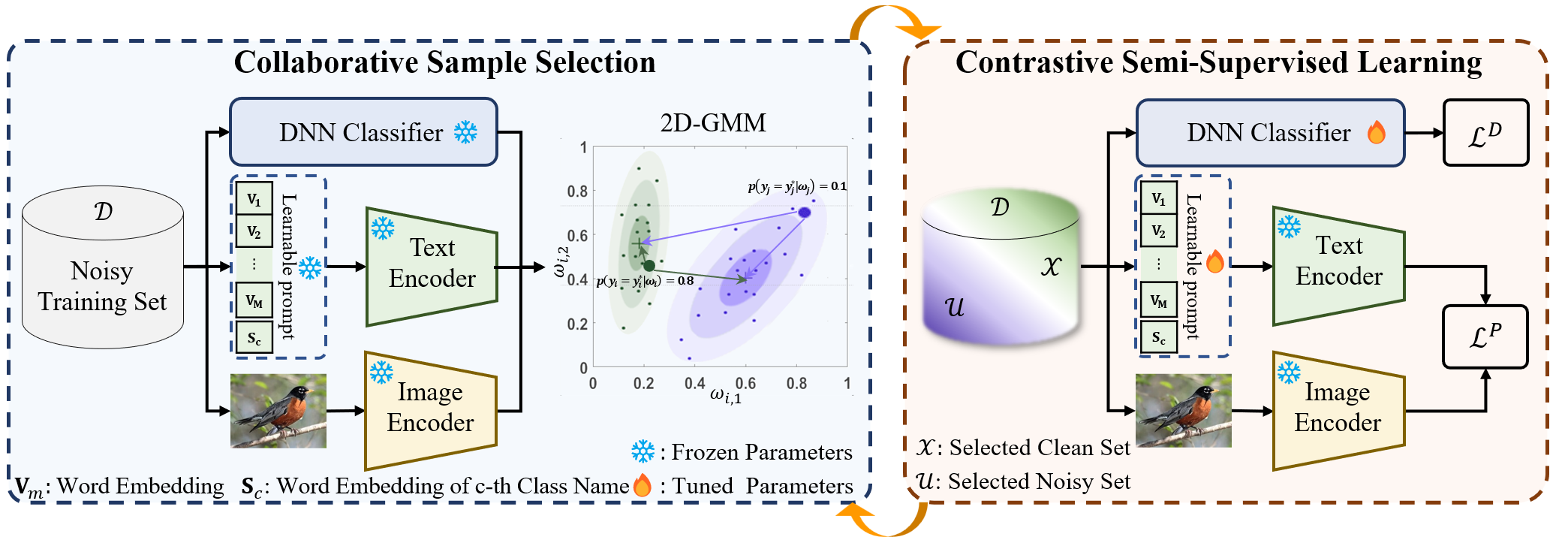}
   \caption{An overview of our alternating framework between Collaborative Sample Selection (CSS) and Contrastive Semi-Supervised Learning (SSL).
   Specifically, the CSS stage models the predictions of DNNs Classifier and CLIP as 2D-GMM to generate the clean probability for data division.
   In the SSL stage, we introduce the contrastive loss to enhance the feature representation, and adopt the co-training mechanism to learn the prompt of CLIP and the parameters of DNNs classifier jointly. It not only improves the indispensable adaptation of CLIP and generalization performance of DNNs, but also benefits to the CSS stage conversely in the next iteration.}
   \label{fig:2}
\end{figure*}

\section{Related Work}
\label{sec:related}
In this section, we provide a brief review of relevant works on learning
with noisy labels, prompt fine-tuning mechanism of the pre-trained model, and self-learning  with unsupervised contrastive loss.
\subsection{Learning with noisy label}
There is an increasing number of studies on learning with noisy labels aimed at alleviating the negative influence of training with noisy labeled samples.
Referring to~\citep{song2022learning}, the survey of prior works is a detailed description of learning with label noise.
The early methods focus on robust loss function for learning with noisy labels.
Several approaches aim to estimate the noise transition matrix~\citep{sukhbaatar2014training,chang2017active,goldberger2016training,lyu2019curriculum}.
However, accurately estimating this matrix proves difficult and may not be practical in real-world scenarios.
Several methods aim to develop noise-tolerant loss functions~\citep{ghosh2017robust,zhang2018generalized}.
For instance, \citep{ghosh2017robust} employs the mean absolute error (MAE) loss, which demonstrates improved generalization compared to the cross-entropy (CE) loss. However, it struggles when confronted with complex datasets.
%

%
In addition, an alternative approach to tackle the challenge of learning with unreliable labels involves employing alternating training: $(i)$ sample selection, $(ii)$ semi-supervised learning. 
The key part of these approaches~\citep{jiang2018mentornet,han2018co,li2020dividemix,karim2022unicon,propmix,sharma2020noiserank,li2022selective,ortego2021multi,jiang2022sparse} relies on the separation of clean samples from the noisy dataset.
\citep{jiang2018mentornet,han2018co,li2020dividemix,karim2022unicon,propmix} adopt the metric on self-generated loss and predictions of the DNN classifiers for selecting the clean samples.
For instance, Co-teaching~\citep{han2018co} trains two networks and the clean samples are identified using the small-loss for updating the parameters of the other network.
Unicon~\citep{karim2022unicon} uniformly selects the clean set using Jensen-Shannon divergence loss.
\citep{jiang2022sparse} proposes a sparse network for distinguishing the clean samples from the small-loss.

Other approaches~\citep{sharma2020noiserank,li2022selective,ortego2021multi,ma2021learning} rely on the separation of clean samples using the feature representations extracted by the classifier. 
However, the DNNs classifier always overfits the noisy labeled samples in the semi-supervised stage especially when the level of label noise is extremely high.
The self-generated information e.g. loss, predictions or feature representations are consistently unable to accurately identify the clean samples.
According to the alternating process between sample selection and semi-supervised learning, the confirmation bias is introduced.
Ultimately, it leads to deteriorates the performance of the classifier.

To handle the problem of the models over-fitting to noisy labeled samples 
during the alternating learning and the confirmation bias in the sample selection stage, recent studies~\citep{karim2022unicon,li2022selective,propmix,yao2021jo} introduce self-supervised learning with the contrastive loss for pre-training the networks.
In unsupervised contrastive learning, DNNs are trained with only images irrespective of the quality of labels for enhancing the feature representation of the models.
However, due to the lack of annotations and the limitation of the amount of training data, the performance of pre-trained classifiers is still unsatisfactory with self-supervised contrastive learning. 
In the sample selection stage, there are still a lot of noisy labeled samples in the clean set that are detected by
the self-generated information of pre-trained models with self-supervised contrastive learning.

\subsection{Prompt tuning}

The recent advancements in large-scale pre-trained vision-language models, such as CLIP~\citep{radford2021learning} and ALIGN ~\citep{jia2021scaling}, have exhibited remarkable potential in acquiring comprehensive representations on massive image-text pairs using contrastive loss.
It allows zero-shot transfer to a variety of downstream classification tasks by matching images and text features and achieves comparable performance.
To make the pre-trained model applicable to downstream tasks, the most popular way to fine-tune CLIP is tuning some learnable prompt of CLIP, such as CoOp~\citep{zhou2022learning} and CoCoOp~\citep{zhou2022conditional}.
The learnable prompt is trained by few-shot learning in CoOp~\citep{zhou2022learning} and CoCoOp~\citep{zhou2022conditional}.
In this paper, we introduce the auxiliary information of the CLIP model as a new source of knowledge to solve the confirmation bias caused by the inaccurate identification of clean labeled samples.
Moreover, to improve the performance and adaptability of CLIP, a co-training mechanism is employed for fine-tuning the prompt of CLIP and training the DNNs classifier jointly using the SSL method with the selected clean and noisy set.
%




\subsection{Contrastive learning}
Contrastive learning methods~\citep{chen2020simple,grill2020bootstrap} aim to learn the robust feature representation by pushing features of positive pairs and spreading features of negative pairs.
In unsupervised learning, the positive pairs are generated by the data augmentation of the same image and the negative pairs are from different images.
It has demonstrated the potential of self-learning with the unsupervised contrastive loss for representation
learning in a large number of researches.
Hence, contrastive learning loss without labels is employed to alleviate the lack of correct annotations in learning with noisy label tasks~\citep{karim2022unicon,li2022selective,propmix,yao2021jo} for pre-training the models.
In this paper, we introduce contrastive loss to improve the ability of feature representation and to prevent the model overfit to noisy samples during the SSL stage.

\section{Proposed Method}
\label{sec:method}
We begin with an overview of the alternating framework between clean sample selection and semi-supervised learning (SSL), which is deployed with the proposed Collaborative Sample Selection (CSS) method in cooperation with the pre-trained model, \ie, CLIP.
Then, we describe the Collaborative Sample Selection method in detail.
Finally, we present the co-training mechanism for the prompt fine-tuning of CLIP and the training of the DNNs classifier in the semi-supervised learning stage. 
\subsection{Overview}
Given a training set $\mathcal{D}=\{(\mathbf{x}_i,\mathbf{y}_i)\}_{i=1}^{N}$ with $N$-sample and $C$-class, in which $\mathbf{x}_i$ denotes the $i$-th training image, and $\mathbf{y}_i \in \{0,1\}^C$ is the annotated label. Correspondingly, we represent the true label of $\mathbf{x}_i$ as $\mathbf{y}_i^{*}$. For real-world scenarios where label noise is inevitable, the annotated label might be randomly corrupted (\ie, $\mathbf{y}_i \neq \mathbf{y}_i^{*}$).
To combat the label noise issue, a popular solution is to adopt the alternating process: $(i)$ sample selection, and $(ii)$ semi-supervised learning. 
In the sample selection stage, a clean subset $\mathcal{X}$ is composed of the samples with correct labels which are identified by the classifier trained using SSL with the noisy training set $\mathcal{D}$.
Accordingly, the remaining samples then constitute a noisy subset $\mathcal{U}$.
The DNNs classifier, which can be regarded as a combination of a feature extractor $f(\cdot)$ parameterized with $\theta_f$ and a classifier $h(\cdot)$ with the parameter $\theta_h$, is then trained by semi-supervised learning. 
However, owing to the self-training manner of DNNs classifier with SSL, some noisy samples are inevitably divided into the clean set in the sample selection stage.
Hence, the confirmation bias is induced by the alternating process and impairs the generalization performance of the DNNs.
Therefore, it is of vital importance to ensure the selected clean subset $\mathcal{X}$ as clean as possible.
%
%
To this end, in this paper, we introduce the large-scale pre-trained model, \ie, CLIP, to assist in clean sample selection. 
By cooperating with the predictions from CLIP and the traditional DNNs classifier, the correctly labeled samples can be accurately distinguished and assigned to the clean subset $\mathcal{X}$.
CLIP~\citep{radford2021learning} is well-known for training with a large amount of open domain data, which makes it powerful in representation. 
Recently, the most popular way to fine-tune CLIP is fine-tuning some learnable prompt of CLIP, such as CoOp~\citep{zhou2022learning} and CoCoOp~\citep{zhou2022conditional}.
In order to improve its performance and adaptability for the task of learning with noisy labels, we employ a co-training mechanism, that performs prompt fine-tuning of CLIP and training of the traditional classification DNNs jointly in semi-supervised learning.
Specifically, the prompt of CLIP is updated using the divided clean and noisy subsets in the SSL stage, as displayed in Figure~\ref{fig:2}.
In addition, we also include the contrastive loss in SSL for improving the feature representation.
In the subsequent sections, we will present comprehensive descriptions of the collaborative sample selection in cooperation with CLIP, and the contrastive semi-supervised co-training mechanism for prompt fine-tuning and DNNs classifier learning.

\subsection{Collaborative Sample Selection with CLIP}
It has been observed that DNNs fit clean samples faster than noisy ones~\citep{liu2020early} during supervised training.
This has inspired many researchers to model cross-entropy loss distribution for clean sample selection~\citep{han2018co,li2020dividemix,propmix} in the early learning phase.
For the training dataset $\mathcal{D}$ with label noise, the cross-entropy (CE) loss for each sample can be formulated as follows:
\begin{equation}
  \ell_i = {\rm CE}(\mathbf{y}_i, \hat{\mathbf{y}}_i) = -\sum_{c=1}^C y_i^c log(\hat{y}_i^c) 
  \label{eq:CE loss}
\end{equation}
where $\hat{\mathbf{y}}_i = h(f(\mathbf{x}_i))$ is the prediction of the image $\mathbf{x}_i$.
Previous state-of-the-art works~\citep{li2020dividemix,propmix} usually fit these loss values by employing a two-component Gaussian Mixture Model (GMM)~\citep{permuter2006study}. The probability of one sample belonging to the clean subset (clean probability), i.e., 
$p(\mathbf{y}_i=\mathbf{y}_i^{*}|\ell_i)$ is then estimated according to the Gaussian component with a smaller mean.
However, since DNNs utilized to calculate the CE loss have been negatively affected by noisy samples in the semi-supervised training stage, the accuracy of the data partition only depending on the loss value can not be guaranteed, particularly in cases of severe label noise.
In this paper, we propose a collaborative sample selection method in cooperation with the pre-trained vision-language model, \ie, CLIP, to ensure the divided clean subset as clean as possible.
CLIP provides two pre-trained encoders, an image encoder $g_{\rm I}(\cdot)$ with the parameter $\theta_1$ and a text encoder $g_{\rm T}(\cdot)$ with the parameter $\theta_2$.
The image encoder is responsible for extracting meaningful feature representations from the input image.
And the text encoder takes the prompts $\mathbf{T}=\! \{\boldsymbol{t}_c\}_{c=1}^C$ as input to generate the classification weight vectors representing the visual concept, where $C$ represents the number of categories.
In CLIP, the prompt $ \boldsymbol{t}_c$ is designed as the word embedding of "a photo of a [CLASS]." where [CLASS] is replaced by the label name, such as "airplane", "bird" or "deer".
For sample $\mathbf{x}_i$, its prediction of CLIP model can be computed by: 
\begin{equation}
  p_i^{CLIP} = \frac{{\rm {exp}}(<g_{\rm I}(\mathbf{x}_i),g_{\rm T}(\boldsymbol{t}_c)>/\tau)}{\sum_{j=1}^C {\rm{exp}}(<g_{\rm I}(\mathbf{x}_i),g_{\rm T}(\boldsymbol{t}_j)>/\tau)},
  \label{eq:CLIP probability}
\end{equation}
where $\boldsymbol{t}_c$ is the word embedding depending on the one-hot label $\mathbf{y}_i$, $\tau$ is a temperature parameter and $< \cdot,\cdot >$ denotes cosine similarity.
Let $\omega_{i,1} = \ell_i$ in Eq.~(\ref{eq:CE loss}) and $\omega_{i,2} = p_i^{CLIP}$ in Eq.~(\ref{eq:CLIP probability}), we have a two-dimensional probability vector ${\boldsymbol{{\omega}}}_i =[\omega_{i,1}, \omega_{i,2}] \in \mathbb{R}^2$. To integrate the prediction from CLIP into the traditional CE loss based sample selection method, we intuitively employ the two-component 2D-GMM to fit $\boldsymbol{\omega}_i$~\citep{d2021consistency}. 
The probability density function (pdf) of the mixture model is then defined as,
\begin{equation}
p(\boldsymbol{\omega}_i) = \sum_{k=1}^K \alpha_k p(\boldsymbol{\omega}_i|k)
\end{equation}
with $K\!=\!2$, and $\alpha_k$ is the weight for each individual 2D-Gaussian pdf $p(\boldsymbol{\omega}_i|k)$ as shown bellow,
\begin{equation}
  p(\boldsymbol{\omega}_i|k) \!=\! \frac{1}{2\pi|\Sigma|^{\frac{1}{2}}}\exp(\!-\!\frac{(\boldsymbol{\omega}_{i}\!-\!\boldsymbol{\mu}_k)^T\Sigma^{\!-\!1}(\boldsymbol{\omega}_{i}\!-\!\boldsymbol{\mu}_k)}{2})
  \label{eq:GMM probability}
\end{equation}
Particularly, $\boldsymbol{\mu}_k$ and $\Sigma$ denote the mean vector and covariance matrix of 2D-Gaussian, respectively.
To fit the two-component 2D-GMM to $\boldsymbol{\omega}$, we use the Expectation-Maximization algorithm\citep{moon1996expectation}. The overall probability of a sample being clean can be computed as:
\begin{equation}
  p(k\! = \! 1|\boldsymbol{\omega}_i) = p(\mathbf{y}_i\! = \! \mathbf{y}_i^*|\boldsymbol{\omega}_i) \!=\! \frac{p(\mathbf{y}_i\! = \! \mathbf{y}_i^*)p(\boldsymbol{\omega}_i|\mathbf{y}_i\! = \! \mathbf{y}_i^*)}{p(\boldsymbol{\omega}_i)}
  \label{eq:GMM probability_1}
\end{equation}
where $k=1$ corresponds to the 2D-Gaussion component with smaller mean.

%
After achieving the overall clean probability, the clean and noisy subsets, i.e., $\mathcal{X}$ and $\mathcal{U}$, are then divided by:
\begin{equation}
\begin{split}
  \mathcal{X} = \{(\mathbf{x}_i,\mathbf{y}_i) | p(\mathbf{y}_i\! = \! \mathbf{y}_i^*|\boldsymbol{\omega}_i) \geq \epsilon \}, \\
  \mathcal{U} = \{(\mathbf{x}_i,\mathbf{y}_i) | p(\mathbf{y}_i\! = \! \mathbf{y}_i^*|\boldsymbol{\omega}_i) < \epsilon \},
  \end{split}
  \label{eq:clean and noisy}
\end{equation}
where $\epsilon = 0.5$ is the threshold.

\subsection{Co-training of the Prompt and DNN Classifiers}
The large-scale pre-trained model CLIP is trained using 400 million (image, text) pairs collected from the internet, which has been demonstrated to show strong visual understanding capability and good classification performance on downstream tasks. 
In CLIP, the prompts $\mathbf{T}\!=\!\{\boldsymbol{t}_c\}_{c=1}^C$ for $C$ classes are designed as the embeddings of "a photo of a [CLASS].", where [CLASS] is the label name.
However, the fixed prompt is considered to be limited in the downstream tasks.

Recent popular works have explored and demonstrated the advantages of prompt fine-tuning of CLIP, such as CoOp~\citep{zhou2022learning} and CoCoOp~\citep{zhou2022conditional}.
Here, to adapt the pre-trained model CLIP to the task of learning with noisy labels, we propose to simultaneously fine-tune the prompts~\citep{zhou2022learning} of CLIP, \ie, $\mathbf{T}=\! \{\boldsymbol{t}_c\}_{c=1}^C$, in semi-supervised learning. 
Typically, the prompt $\boldsymbol{t}_c = [\mathbf{V}_1,\mathbf{V}_2\ldots \mathbf{V}_m \ldots \mathbf{V}_M ,\mathbf{S}_c]$ includes the learnable features of context tokens, \ie, $\mathbf{V}_m \in \mathbb{R}^{512}$, and word embedding of the $c$-th class name, \ie, $\mathbf{S}_c \in \mathbb{R}^{512}$. The number of context tokens, \ie, $M$, is a hyperparameter and empirically set as 16.
In the implementation, compared random initialization with manual initialization, we employ the embeddings of “a photo of a” to initialize the context tokens.

%

%

%
In the semi-supervised learning stage,  we propose to jointly fine-tune the learnable word embeddings $\mathbf{V}_m$ of the prompt and train the DNNs classifier following FixMatch~\citep{sohn2020fixmatch}.
For the labeled sample $\mathbf{x}_i$ in the divided clean subset $\mathcal{X}=\{(\mathbf{x}_i,\mathbf{y}_i)\}_{i=1}^{N_1}$, where $N_1$ is the number of clean labeled samples, a supervised cross-entropy loss $\mathcal{L}_{\mathcal{X}}$ is adopted on its weakly-augmented version $\mathbf{x}^w_i$ to optimize the DNNs classifier, as follows:
\begin{equation}
 \mathcal{L}_{\mathcal{X}}^{\rm D} = \frac{1}{N_1}\sum_{i=1}^{N_1} {\rm CE}(\mathbf{y}_i, h(f(\mathbf{x}^w_i))).
  \label{eq:label loss}
\end{equation}
For fine-tuning the prompt, the objective with labeled samples is formulated as:
\begin{equation}
 \mathcal{L}_{\mathcal{X}}^{\rm P} = \frac{1}{N_1}\sum_{i=1}^{N_1} {\rm CE}(\mathbf{y}_i, \mathbf{p}^{CLIP}(\mathbf{x}_i,\mathbf{T})),
  \label{eq:label loss of prompt}
\end{equation}
where $\mathbf{p}^{CLIP}(\mathbf{x}_i,\mathbf{T})=\{p_{c}^{CLIP}(\mathbf{x}_i,\mathbf{t_c})\}^{C}_{c=1}$ is the prediction of CLIP model calculated as follows: 
\begin{equation}
  p_c^{CLIP}(\mathbf{x}_i,\mathbf{t_c}) = \frac{{\rm {exp}}(<g_{\rm I}(\mathbf{x}_i),g_{\rm T}(\boldsymbol{t}_c)>/\tau)}{\sum_{j=1}^C {\rm{exp}}(<g_{\rm I}(\mathbf{x}_i),g_{\rm T}(\boldsymbol{t}_j)>/\tau)}.
  \label{eq:CLIP prediction}
\end{equation}

For the unlabeled example $\mathbf{u}_i$ in subset $\mathcal{U}=\{\mathbf{u}_i\}_{i=1}^{N_2}$, we weakly augment it to obtain $\mathbf{u}^w_i$, and forward the augmented version to achieve the prediction as the pseudo-label.
To train the DNNs classifier, the pseudo-label is calculated by $\hat{\mathbf{y}}_i^{\rm D}= h(f(\mathbf{u}^w_i))$.
To fine-tune the prompt, the pseudo-label is calculated as $\tilde{\mathbf{y}}_i^{\rm P}=\{p_c^{CLIP}(\mathbf{u}^w_i,\mathbf{t_c})\}^{C}_{c=1}$ according to Eq.~(\ref{eq:CLIP prediction}).
%

%

Subsequently, we utilize cross-entropy loss to enforce the model output for its strongly-augmented version $\mathbf{u}^s_i$ to be consistent with the pseudo-label whose maximum probability surpasses a predefined threshold value $\delta$.
Therefore, the unsupervised loss on $\mathcal{U}$ for optimizing DNNs is formulated as:
\begin{equation}
	\mathcal{L}_{\mathcal{U}}^{\rm D} = \frac{1}{N_2}\sum_{i=1}^{N_2} \mathbb{1}(\max (\hat{\mathbf{y}}_i^{\rm D}) > \delta) {\rm CE} (\hat{\mathbf{y}}_i^{\rm D}, h(f(\mathbf{u}^{s}_i))).
\end{equation}
where $\max(\cdot)$ gets the maximum value and $\delta=0.95$ is the threshold.
In the same way, the unsupervised loss for fine-tuning the prompt is defined as:
\begin{equation}
	\mathcal{L}_{\mathcal{U}}^{\rm P} = \frac{1}{N_2}\sum_{i=1}^{N_2} \mathbb{1}(\max(\tilde{\mathbf{y}}_i^{\rm P}) > \delta) {\rm CE} (\tilde{\mathbf{y}}_i^{\rm P},\mathbf{p}^{CLIP}(\mathbf{u}^{s}_i,\mathbf{T})).
\end{equation}
where $\mathbf{p}^{CLIP}(\mathbf{u}_i^s,\mathbf{T})=\{p_{c}^{CLIP}(\mathbf{u}^{s}_i,\mathbf{t_c})\}^{C}_{c=1}$ is the prediction of CLIP model similar as Eq.~(\ref{eq:CLIP prediction}) with inputs $\mathbf{u}^{s}_i$ and $\mathbf{t_c}$.
%

%
To enhance the feature representation, we also introduce the contrastive loss in SSL, which pushes the feature vectors ($f(\mathbf{x}_i^{w_1})$ and $f(\mathbf{x}_i^{w_2})$) from two views (weakly-augmented versions) of the same image $\mathbf{x}_i$ and spreading features of negative pairs, as following:
\begin{equation}
	\mathcal{L}_{con}  = -\frac{1}{N}\sum_{i=1}^{N} \log\frac{\exp{(<f(\mathbf{x}_i^{w_1}),f(\mathbf{x}_i^{w_2})>/\tau)}}{\sum_{j=1}^{2N}\mathbb{1}_{j \neq i} \exp(<f(\mathbf{x}^w_i), f(\mathbf{x}^w_j)>/\tau)},
\end{equation}
where $\tau$ is a temperature parameter and $< \cdot,\cdot >$ denotes cosine similarity.

To mitigate the issue of predicting all samples to the same class, we are currently adopting the methodology presented in DivideMix~\citep{li2020dividemix} and  integrating the regularization technique $\mathcal{L}_{reg}$ into our model.
$\mathcal{L}_{reg}$ utilizes a uniform prior distribution to effectively regularize the average model output across all training samples:
\begin{equation}
	\mathcal{L}_{reg}  = \sum_{c=1}^C \pi_c \log \bigg (\pi_c / \frac{\sum_{\mathbf{x} \in  \mathcal{X} + \mathcal{U}} \mathbf{o}_c }{|\mathcal{X}|+|\mathcal{U}|} \bigg),
\end{equation}
where $\mathbf{o} = h(f(\mathbf{x}))$ represents the output of the DNN classifier and $\mathbf{o}_c$ denoting the predicted probability for class $c$ and $\pi_c = 1/C$.
Overall, to update the parameters of DNNs, \ie, $\theta_f$, $\theta_h$, we aim to solve the following minimization problem with the total loss:
\begin{equation}
 \mathop{\arg \min}\limits_{\theta_f,\theta_h} (	\mathcal{L}^{\rm D} = \mathcal{L}_{\mathcal{X}}^{\rm D} \! +\! \lambda_{u} \mathcal{L}_{\mathcal{U}}^{\rm D} \! + \!\lambda_{c} \mathcal{L}_{con}  \!+\!\lambda_{r} \mathcal{L}_{reg}),
 \label{eq:DNNs}
\end{equation}
 where $\lambda_{u}$, $\lambda_{c}$, and $\lambda_{r}$ are the hyperparameters for controling the strength of the unsupervised loss, the contrastive loss, and the regularization term, and $\lambda_{u}$, $\lambda_{c}$, and $\lambda_{r}$ are set as $0.5$, $0.025$, and $1$ in our experiments, respectively.


%
%
%

%
To fine-tune the prompt, we update $\mathbf{V}_m$ in $\mathbf{t}_c$ in semi-supervised learning with the following objective:
\begin{equation}
   \mathop{\arg\min}\limits_{\mathbf{V}_m}({\mathcal{L}}^{\rm P} =\mathcal{L}_{\mathcal{X}}^{ \rm P} + \lambda_{u} \mathcal{L}_{\mathcal{U}}^{\rm P}).
    \label{eq:prompt}
\end{equation}
Finally, by alternately selecting the clean labeled sample and updating the DNNs and fine-tuning the prompt of the CLIP model at each epoch, it fulfills a positive cycle that a cleaner set will result in a better representation of DNNs and recognition ability of CLIP, which in turn, will benefit to select cleaner labeled samples. Algorithm 1 delineates the full algorithm.
%


\begin{algorithm}[!htp]
    \footnotesize
    \setstretch{1.1}
    \caption{Collaborative Sample Selection (CSS)}
    \begin{algorithmic}[1] 
        \REQUIRE Training set $\mathcal{D}=\{(\mathbf{x}_i,\mathbf{y}_i)\}_{i=1}^{N}$, number of samples $N$, number of labeled samples $N_1$, number of unlabeled samples $N_2$, prompt $\boldsymbol{t}_c = [\mathbf{V}_1,\mathbf{V}_2\ldots \mathbf{V}_m \ldots \mathbf{V}_M ,\mathbf{S}_c]$, feature extractor $f(\cdot)$ with $\theta_f$, classifier $h(\cdot)$ with $\theta_h$, image encoder $g_{I}(\cdot)$ and text encoder $g_{T}(\cdot)$ of CLIP, temperature parameter $\tau$, threshold $\delta$ and $\epsilon$, unsupervised loss weight $\lambda_{u}$, contrastive loss weight $\lambda_{c}$, regularization term weight $\lambda_{r}$, $\pi_c = 1/C$
        \ENSURE Prediction of classifier $\hat{\mathbf{y}}_i$
        \STATE {$\theta_f$ = Pre-trained$(f(\mathcal{D}))$  \footnotesize //self-supervised pre-training with contrastive loss}
        \STATE {$\theta_f, \theta_h =$ Warm\_up($\mathcal{D},\theta_f,\theta_h)$ }
        \WHILE{$e < $MaxEpoch}
            \STATE {\footnotesize //collaborative sample selection}
            \FOR {$ i = \{1,\cdots,|\mathcal{D}|\}$} 
                \STATE { $\hat{\mathbf{y}}_i = h(f(\mathbf{x}_i))$}
                \STATE {Estimate $\omega_{i,1} = \ell_i,$ with $\ell_i = -\sum_{c=1}^C y_i^c log(\hat{y}_i^c)$}
                \STATE {Estimate  $p_i^{CLIP} = \frac{{\rm {exp}}(<g_{\rm I}(\mathbf{x}_i),g_{\rm T}(\boldsymbol{t}_c)>/\tau)}{\sum_{j=1}^C {\rm{exp}}(<g_{\rm I}(\mathbf{x}_i),g_{\rm T}(\boldsymbol{t}_j)>/\tau)}$}
                \STATE{$\omega_{i,2} = p_i^{CLIP}$}
               \STATE {$\boldsymbol{\omega}_i =[\omega_{i,1}, \omega_{i,2}]$}
            \ENDFOR
            \STATE{$ p(\mathbf{y}_i\! = \! \mathbf{y}_i^*|\boldsymbol{\omega}_i) = $ 2D-GMM $(\boldsymbol{\omega}_i)$}
            \STATE{$\mathcal{X}\!=\! \{(\mathbf{x}_i,\mathbf{y}_i)|  p(\mathbf{y}_i\! = \! \mathbf{y}_i^*|\boldsymbol{\omega}_i) \!\geq\! \epsilon \}$}
            \STATE{$\mathcal{U}\!=\!\{(\mathbf{x}_i,\mathbf{y}_i)|p(\mathbf{y}_i\! = \! \mathbf{y}_i^*|\boldsymbol{\omega}_i) \!<\!\epsilon\}$}
            \STATE {\footnotesize//contrastive semi-supervised co-training}
            \FOR {$ i = \{1,\cdots,|\mathcal{D}|\}$}
            \STATE {$\mathbf{x}^w_i$= WeakAugment($\mathbf{x}_i$), $\mathbf{u}^w_i$= WeakAugment($\mathbf{u}_i$)}
            \STATE {$\mathbf{u}^s_i$= StrongAugment($\mathbf{u}_i$)}
            \STATE {$\hat{\mathbf{y}}_i^{\rm D}= h(f(\mathbf{u}^w_i))$}
            
           \STATE { $\tilde{\mathbf{y}}_i^{\rm P}=\{p_c^{CLIP}(\mathbf{u}^w_i,\mathbf{t_c})\}^{C}_{c=1}$}
            \ENDFOR
            \STATE {$\mathcal{L}_{\mathcal{X}}^{\rm D} = \frac{1}{N_1}\sum_{i=1}^{N_1} {\rm CE}(\mathbf{y}_i, h(f(\mathbf{x}^w_i)))$}
     
            \STATE {$  \mathcal{L}_{\mathcal{X}}^{\rm P} = \frac{1}{N_1}\sum_{i=1}^{N_1} {\rm CE}(\mathbf{y}_i, \mathbf{p}^{CLIP}(\mathbf{x}_i,\mathbf{T}))$}

            \STATE {$\mathcal{L}_{\mathcal{U}}^{\rm D} = \frac{1}{N_2}\sum_{i=1}^{N_2} \mathbb{1}((\hat{\mathbf{y}}_i^{\rm D})^{max} > \delta) {\rm CE} (\hat{\mathbf{y}}_i^{\rm D}, h(f(\mathbf{u}^{s}_i)))$}
            
            \STATE {$			\mathcal{L}_{\mathcal{U}}^{\rm P} = \frac{1}{N_2}\sum_{i=1}^{N_2} \mathbb{1}(\max(\tilde{\mathbf{y}}_i^{\rm P}) > \delta) {\rm CE} (\tilde{\mathbf{y}}_i^{\rm P},\mathbf{p}^{CLIP}(\mathbf{u}^{s}_i,\mathbf{T}))$}
            \STATE {\vspace{0.25em}
            $\mathcal{L}_{con}\!=\! -\frac{1}{N}\sum_{i=1}^{N} \log\frac{\exp{(<f(\mathbf{x}_i^{w_1}),f(\mathbf{x}_i^{w_2})>/\tau)}}{\sum_{j=1}^{2N}\mathbb{1}_{j \neq i} \exp(<f(\mathbf{x}^w_i), f(\mathbf{x}^w_j)>/\tau)}$ \vspace{0.25em}
            }
            \STATE{	$\mathcal{L}_{reg}\!=\!\sum_{c=1}^C \pi_c \log \bigg (\pi_c / \frac{\sum_{\mathbf{x} \in  \mathcal{X} + \mathcal{U}} \mathbf{o}_c }{|\mathcal{X}|+|\mathcal{U}|} \bigg)$ \footnotesize //where $\mathbf{o}\!=\!h(f(\mathbf{x}))$}
            \STATE{$\mathcal{L}^{\rm D} = \mathcal{L}_{\mathcal{X}}^{\rm D} \! +\! \lambda_{u} \mathcal{L}_{\mathcal{U}}^{\rm D} \! + \!\lambda_{c} \mathcal{L}_{con}  \!+\!\lambda_{r} \mathcal{L}_{reg}$}
            \STATE{${\mathcal{L}}^{\rm P} =\mathcal{L}_{\mathcal{X}}^{ \rm P} + \lambda_{u} \mathcal{L}_{\mathcal{U}}^{\rm P}$}
            \STATE{Update $\theta_f, \theta_h$ with $\mathcal{L}^{\rm D}$ in Eq.~(\ref{eq:DNNs})}
            \STATE{Update $\mathbf{V}_m$ with $\mathcal{L}^{\rm P}$  in Eq.~(\ref{eq:prompt})}
        \ENDWHILE
    \end{algorithmic} \end{algorithm}

\section{Experiments}
In this section, we conducted comprehensive experiments on both synthetic and real-world label noise to validate the efficacy of our method.
We first introduce the details of the experiments.
Then, we show the experimental results compared with the state-of-the-art methods on several benchmark datasets at multiple noise levels.
Finally, we present a number of ablation studies to provide more insights and analysis of the proposed method in detail.

\subsection{Datasets and implementation details}

\noindent \textbf{Datasets.}  For evaluating the synthetic label noise, we conducted experiments on the CIFAR-10~\citep{khosla2020supervised} and CIFAR-100~\citep{krizhevsky2009learning}  datasets. 
We assess the impact of real-world noise by conducting evaluations on the Clothing1M~\citep{xiao2015learning} and WebVision~\citep{li2017webvision} datasets.
The CIFAR-10 dataset consists of $50,000$ training images and $10,000$ test images, each with a size of $32\times32\times3$ pixels, divided into 10 distinct classes.
CIFAR-100 is like the CIFAR-10, except with 100 categories.
Both Clothing1M and WebVision are significant large-scale datasets that are known for their incorporation of real-world label noise. These datasets serve as valuable resources for studying and addressing the challenges posed by noisy labels in practical settings.
Clothing1M offers an extensive dataset with an astounding one million training images and 10,000 test images, all focused on 14 diverse cloth-related classes. These images were thoughtfully collected from online shopping websites and are conveniently sized at $256\times 256$ pixels, providing researchers and analysts with a rich and comprehensive resource for their studies.
The WebVision dataset is an extensive collection of 2.4 million images gathered from popular platforms such as Flickr and Google. These images are associated with the same $1000$ classes as the ILSVRC12 dataset~\citep{deng2009imagenet} and have been uniformly resized to dimensions of 256×256 pixels. To ensure consistency with previous studies~\citep{li2020dividemix, propmix}, our analysis focuses specifically on the initial $50$ classes from the Google image subset.

Following prior works~\citep{li2020dividemix,arazo2019unsupervised,jiang2020beyond}, We conduct evaluations on the CIFAR-10 and CIFAR-100 datasets to analyze the effects of synthetic label noise. Specifically, we evaluate two types of label noise: \textit{symmetric} and \textit{asymmetric}. By examining the impact of these noise types, we gain insights into their influence on the performance of models trained on these datasets.
Symmetric noise refers to a proportion of training samples with their labels are uniformly distributed to all possible classes.
In experiments, the proportions are 20\%, 50\%, 80\%, and 90\% as used in~\citep{propmix,li2020dividemix}.
As for the asymmetric noise, the labels are flipped to the semantically similar classes as a more realistic setting~\citep{patrini2017making}, e.g. deer $\rightarrow$ horse, dog $\leftrightarrow$ cat.
In our study, we investigate the effects of asymmetric label noise at different rates: 10\%, 30\%, and 40\%. 
Under realistic noise, Clothing1M contains a high-level asymmetric label noise since the images are searched by some similar types of clothing on websites and the total noise rate is around 38\%.
While, in WebVision, there are many real-world noisy labels without human annotation due to searches on the website.
\begin{table*}[t]
\scriptsize
\centering
\caption{Comparison with the state-of-the-art methods in terms of test accuracy (\%) on CIFAR-10 and CIFAR-100 with different symmetric and asymmetric noise levels.}
	\label{tb:cifar}
	\begin{tabular}{l|p{0.3cm}<{\centering} p{0.3cm}<{\centering} p{0.3cm}<{\centering} p{0.4cm}<{\centering}|p{0.3cm}<{\centering} p{0.3cm}<{\centering} p{0.4cm}<{\centering}|p{0.3cm}<{\centering} p{0.3cm}<{\centering} p{0.3cm}<{\centering} p{0.4cm}<{\centering} |p{0.3cm}<{\centering} p{0.3cm}<{\centering} p{0.3cm}<{\centering}}
		\hlinew{1.3pt}
		\multicolumn{1}{c|}{Dataset} & \multicolumn{7}{c}{CIFAR-10}  &  \multicolumn{7}{|c}{CIFAR-100} \\
		\hline
		\multicolumn{1}{c|}{Noise type} & \multicolumn{4}{c|}{sym.} & \multicolumn{3}{c|}{asym.}&  \multicolumn{4}{c|}{sym.}& \multicolumn{3}{c}{asym.}\\
		\multicolumn{1}{c|}{Method/Noise Ratio} & 20\% & 50\% & 80\% & 90\% & 10\% & 30\% &40\% & 20\% & 50\% & 80\% & 90\% & 10\%  & 30\% &40\% \\
		\hline
		\multirow{1}{*}{Cross-Entropy}  &86.8 & 79.4 & 62.9 & 42.7 & 88.8& 81.7&85.0 &62.0 & 46.7 & 19.9 & 10.1&68.1& 53.3& 44.5 \\
		\multirow{1}{*}{P-correction~\citep{yi2019probabilistic}}  & 92.0 &88.7& 76.5 &58.2&93.1 & 92.6& 91.6& 68.1 &56.4& 20.7& 8.8 &76.1 &59.3& 48.3\\
	PENCIL~\citep{yi2019probabilistic} & 92.4& 89.1& 77.5& 58.9& 93.1 & 92.9& 88.5& 69.4& 57.5& 31.1 &15.3&76.0& 59.3& 48.3\\
		\multirow{1}{*}{M-correction\citep{arazo2019unsupervised}}& 93.8 & 91.9 & 86.6 & 68.7& 89.6 & 92.2 & 91.2 &73.4 & 65.4& 47.6 & 20.5 & 67.1 & 58.6 & 47.4 \\
		\multirow{1}{*}{ELR~\citep{liu2020early}} & 95.8 & 94.8 & 93.3 & 78.7& 95.4& 94.7 & 93.0 & 77.6 & 73.6& 60.8 & 33.4 &77.4 &75.1 &74.0\\
		\multirow{1}{*}{DivideMix~\citep{li2020dividemix}} & 96.1 & 94.6 & 93.2 & 76.0& 93.8 & 92.5 & 91.7 &77.3 & 74.6& 60.2 & 31.5& 71.6 & 69.5 & 55.1 \\
		\multirow{1}{*}{MOIT~\citep{ortego2021multi}} & 94.1 &  91.1 & 75.8  & 70.1&94.2 &94.1 & 93.2&78.9 & 70.1 & 51.4 & 24.5 &77.4 &75.1 &74.0\\
		\multirow{1}{*}{UNICON~\citep{karim2022unicon}}&  96.0  & 95.6 & 93.9 &  90.8& 95.3 &94.8  & 94.1 & 78.9 & 77.6 & 63.9  & 44.8 & 78.2 &75.6 & 74.8 \\
		\multirow{1}{*}{Sel-CL+~\citep{li2022selective}}
    		& 95.5 & 93.9 & 89.2 & 81.9 &95.6  & 94.5 & 93.4 & 76.5 & 72.4  & 59.6 & 48.8 & 78.7 &77.5 & 74.2  \\
		\multirow{1}{*}{NCR+~\citep{iscen2022learning}}
		 & 95.2& 94.3 & 91.6  & 75.1& - &  - & 90.7 & 76.6 & 72.5  & 58.0 & 30.8 &- & -& -\\
		\multirow{1}{*}{PropMix~\citep{propmix}} & 96.4 & 95.8 & 93.9 & 93.5 &95.7 & 95.0 & 94.9 &77.4 & 74.6 & 67.3 & 58.6 &77.1 & 71.1& 60.2\\
		\hline
		\multirow{1}{*}{Ours}
		& \textbf{96.5} & \textbf{96.3} & \textbf{95.6} & \textbf{94.4} & \textbf{95.8} & \textbf{95.2} & \textbf{95.0}&\textbf{79.1} & \textbf{77.7}& \textbf{72.5} & \textbf{68.7} & \textbf{78.9} &\textbf{77.6} & \textbf{76.1} \\
		\hlinew{1.3pt}
	\end{tabular}
\end{table*}

\noindent \textbf{Implementation details.}
For CIFAR-10 and CIFAR-100, we employ the 18-layer PreAct Resnet~\citep{he2016identity} architecture as our backbone model following previous methods~\citep{li2020dividemix,propmix}.
In the pre-trained stage, the model is trained exclusively using the contrastive loss for $800$ epochs. The goal is to update the parameter $\theta_f$ of the feature extractor. The training is performed with a batch size of $1024$, following the approach outlined in~\citep{propmix}.
Next, we conduct a warm-up training phase on the noisy training dataset, consisting of $10$ epochs for CIFAR-10 and $30$ epochs for CIFAR-100. 
In the warm-up stage, the DNNs classifier (with $\theta_f$ and $\theta_h$) is updated by a standard supervised learning method using cross-entropy loss.
Finally, the network is trained for $300$ epochs, utilizing a batch size of $128$. Stochastic gradient descent (SGD) optimizer is employed for training, with a momentum of $0.9$ and a weight decay of $5e-4$.
The initial learning rate is set to $0.02$ and is subsequently reduced by a factor of $10$ after $150$ epochs.

For Clothing1M, we fine-tune the pre-trained ResNet-50 model for 80 epochs, utilizing a batch size of 32 and only warm up the network for 1 epoch following previous work~\citep{propmix}.
The initial learning rate is $0.001$ and decay after $40$ epochs with a weight decay of $1e-3$.
We take an InceptionResNet-V2
network as the backbone on the real-world dataset WebVision-50.
During the training process, we utilized SGD with a momentum of 0.9, a weight decay of $5e-4$, and an initial learning rate of $0.01$ for a total of $80$ epochs.

\begin{table}[htbp!]
	\centering
	\footnotesize
	\caption{Comparison with the state-of-the-art methods in terms of test accuracy (\%) on Clothing1M and WebVision. The {\color{red}{first}}, {\color{green}{second}} and {\color{blue}{third}} best results are highlighted in different colors.}\label{tb:clothing1M}
	\begin{tabular}{l|c|c c}
		\hlinew{1.3pt}
		\multirow{2}{*}{Methods} $\qquad$  & \multirow{2}{*}{Clothing1M} & \multicolumn{2}{c}{WebVision}\\ $\qquad$  &  $\qquad$ & Top1 & Top5\\
		\hline
          Decoupling~\citep{malach2017decoupling} $ \qquad$ & \makecell[c]{-}& 62.54 & 84.74\\
        Co-teaching~\citep{han2018co} $ \qquad$ & \makecell[c]{-}& 63.58 &  85.20\\
        MentorMix~\citep{jiang2020beyond} $ \qquad$ & \makecell[c]{-}& 76.00 &  90.20\\
        DivideMix~\citep{li2020dividemix} $ \qquad$ & \makecell[c]{74.76}& 77.32 & 91.64\\
	ELR+~\citep{liu2020early} $ \qquad$ & \makecell[c]{\color{blue}{\textbf{74.81}}} & 77.78 & 91.64\\
  SOP~\citep{zhang2020symmetry} $ \qquad$ & \makecell[c]{73.50}& 76.60 & -\\
		PropMix~\citep{propmix} $ \qquad$ & \makecell[c]{74.30}& {\color{blue}\textbf{78.84}} & 90.56\\
		UNICON~\citep{karim2022unicon} $ \qquad$ & \makecell[c]{{\color{green}\textbf{74.98}}}& 77.60 & {\color{red}\textbf{93.44}}\\
		Sel-CL+~\citep{li2022selective} $ \qquad$ & \makecell[c]{-}& {\color{red}{\textbf{79.96}}} &{\color{blue}\textbf{92.64}}\\
		NCR+~\citep{iscen2022learning}$ \qquad$ & \makecell[c]{74.60}& 76.80 & -\\
		CTRR~\citep{yi2022learning}$ \qquad$ & \makecell[c]{74.60}& - & -\\
		\hline
		Ours & \makecell[c]{\textbf{{\color{red}75.09}}}&\makecell[c]{{\color{green}\textbf{79.01}}} &\makecell[c]{\color{green}\textbf{93.08}}\\
		\hlinew{1.3pt}
	\end{tabular}
 \label{tb:clothing and webvision}
\end{table}
\subsection{Comparison with state-of-the-art methods}
\noindent \textbf{CIFAR-10 and CIFAR-100 datasets.} 
In Table~\ref{tb:cifar}, we provide the average test accuracy over the final 10 epochs for CIFAR-10 and CIFAR-100 datasets, considering both symmetric and asymmetric label noise.
For symmetric noise, one can observe that our method achieves substantial improvements in the case of extremely high noise levels on both CIFAR-10 and CIFAR-100 datasets.
On CIFAR-10, our proposed method outperforms the baseline, \ie, PropMix, with around $1\% \sim 2\%$ in terms of accuracy on $80\%$ and $90\%$ noise rates.
%
%
On CIFAR-100, due to the more categories, classification becomes more challenging, especially with extreme label noise.
In Table~\ref{tb:cifar}, when the noise ratio is $90\%$, the test accuracy of most of the state-of-the-art methods on CIFAR-100 can only achieve $58.6\%$.
However, we can obtain an accuracy of $68.7\%$, with a significant gain of $\sim 10.0\%$ over PropMix.
Also at the noise ratio of $80\%$, the proposed approach improves PropMix by $5.2\%$.  
The above results effectively prove that the proposed sample selection method with the assistance of CLIP effectively reduces the over-fitting of label noise.
Moreover, it is noteworthy that our method consistently achieves competitive results, demonstrating its efficacy and strength.
In comparison to the baseline, our proposed method brings around $0.1\% \sim 0.5\%$  and $1\% \sim 3\%$  at low noise ratios (20\% and 50\%) on CIFAR-10 and CIFAR-100, respectively. 
For asymmetric noise, we report the test accuracy on CIFAR-10 and CIFAR-100 in Table~\ref{tb:cifar}.
It is worth noting that our approach significantly outperforms traditional baselines and performs favorably compared to the state-of-the-art methods (around $0.1\% \sim 1.3\%$) at all noise ratios.
The results indicate that our method excels in achieving the best performance with more realistic noise.
\begin{table*}[htbp!]
\small
\centering
\caption{Ablation study to evaluate the effect of each component on CIFAR-100 in terms of test accuracy (\%).}
 \begin{tabular}{p{2.8cm}<{\centering} p{1.8cm}<{\centering}p{2.5cm}<{\centering}|cccc}
    \hlinew{1.3pt}

    \multirow{2}{*}{\makecell[c]{Collaborative \\ sample selection}}& 
    
    \multirow{2}{*}{\makecell[c]{Prompt \\ fine-tuning}}& 
    
    \multirow{2}{*}{\makecell[c]{Contrastive \\ loss ($\mathcal{L}_{con}$)}}
    & \multicolumn{4}{c}{CIFAR-100}\\
    \makebox[0.00001\textwidth][c]{}& & 
    & \makebox[0.06\textwidth][c]{20\%} 
    & \makebox[0.06\textwidth][c]{50\%}
    & \makebox[0.06\textwidth][c]{80\%}
    & \makebox[0.06\textwidth][c]{90\%}\\
   \hline
    \ding{55}&\ding{55}& \ding{55}& 
    \makebox[0.06\textwidth][c]{78.34} & \makebox[0.06\textwidth][c]{74.48} & \makebox[0.06\textwidth][c]{67.21} & \makebox[0.06\textwidth][c]{58.08}\\
    \ding{55}& \ding{55} & $\checkmark$& 
    \makebox[0.06\textwidth][c]{78.80} & \makebox[0.06\textwidth][c]{74.64} & \makebox[0.06\textwidth][c]{67.92} & \makebox[0.06\textwidth][c]{58.72}\\    
    $\checkmark$ &\ding{55} &$\checkmark$ &
     \makebox[0.06\textwidth][c]{78.81} & \makebox[0.06\textwidth][c]{75.13} & \makebox[0.06\textwidth][c]{71.29} &\makebox[0.06\textwidth][c]{66.09} \\

  $\checkmark$& $\checkmark$ &$\checkmark$ &
    \textbf{79.12} & \textbf{77.65}& \textbf{72.46} & \textbf{68.66}\\
    \hlinew{1.pt}
    \multicolumn{4}{l}{\footnotesize*Zero-shot CLIP test accuracy$=63.71\%$.}
\end{tabular}
\label{tb:ablation1}
\end{table*}

\noindent \textbf{Clothing1M dataset.} 
In Table~\ref{tb:clothing and webvision}, we evaluate our proposed approach against state-of-the-art methods on a real-world dataset Clothing1M.
In comparison with the baseline method PropMix~\citep{propmix}, we achieve a test accuracy of 75.09\%, with a slight improvement of 0.7\%. 
Moreover, our method not only demonstrates favorable performance but also outperforms recent state-of-the-art methods.
The results unequivocally demonstrate the exceptional effectiveness of our proposed method on real-world noise datasets, particularly in the presence of complex label noise.
\noindent \textbf{WebVision dataset.} 
We also evaluate on the popular large-scale real dataset WebVision~\citep{li2017webvision}, which contains various types of label noise.
We provide the top-1 and top-5 accuracy metrics on the WebVision validation set as part of our evaluation and mark them with different colors in Table~\ref{tb:clothing1M}.
Our method consistently surpasses the baseline in terms of both top-1 and top-5 accuracy and obtains around 2.5\% improvement in top-5 accuracy compared with PropMix.
We also achieve competitive performance compared to the other state-of-the-art methods.
Even though WebVision contains a large number of images and various types of label noise, our method still has favorable performance.
\subsection{Ablation study}
We conduct an ablation study on CIFAR-100 dataset with symmetric label noise at multiple rates to gain insights into the effectiveness of our proposed method.
\noindent \textbf{Effect of each component.}
The proposed method mainly contains three contributions: contrastive loss in SSL, collaborative sample selection with CLIP, and prompt fine-tuning.
Here, we experimentally verify the effectiveness of each component in our proposed method.
We design our baseline approach following the framework of PropMix for a fair and clear comparison.
Specifically, we also begin with self-supervised pre-training using the contrastive loss to initialize the feature extractor of DNNs, \ie, $f(\cdot)$ parameterized with $\theta_f$.
Then, we warm up the classifier $h(\cdot)$  with the parameter $\theta_h$ using the cross-entropy loss with the noisy training set.
When processing sample selection and semi-supervised learning alternately and iteratively, unlike PropMix, we use simple GMM to divide the training set as in DivideMix~\citep{li2020dividemix}, and adopt FixMatch~\citep{sohn2020fixmatch} for the SSL phase.
The first row in Table~\ref{tb:ablation1} gives the results of our baseline. 

On the basis of the baseline method, we introduce the contrastive loss $\mathcal{L}_{con}$ during semi-supervised learning to constrain the feature representations.
As shown in Table~\ref{tb:ablation1} (second row), by adding the contrastive loss in the SSL stage, the classification performance is improved with an accuracy gain of $ 0.16\% \sim 0.76\%$ under different noise ratios.
It demonstrates that self-learning by pushing the features of positive pairs closer together while spreading the features of negative pairs further apart is effective for semi-supervised training.
%

%
%
\begin{figure}[htbp!]
\hspace{-1.5em}
\subfloat{
\includegraphics[width=0.53\linewidth]{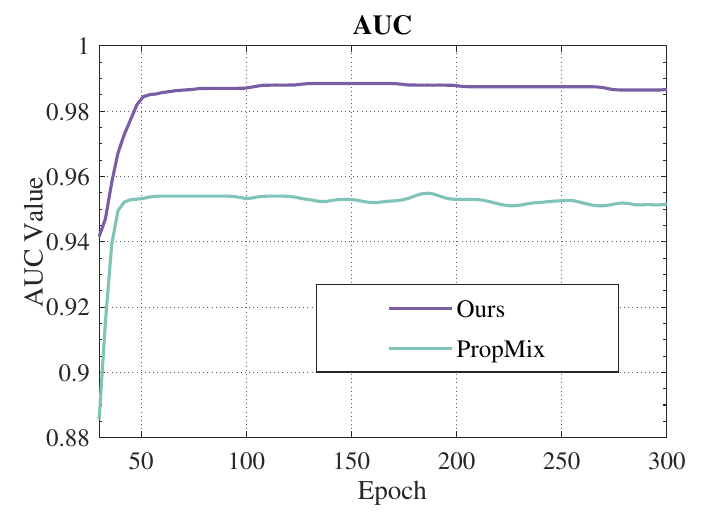}}
\subfloat{
\includegraphics[width=0.53\linewidth]{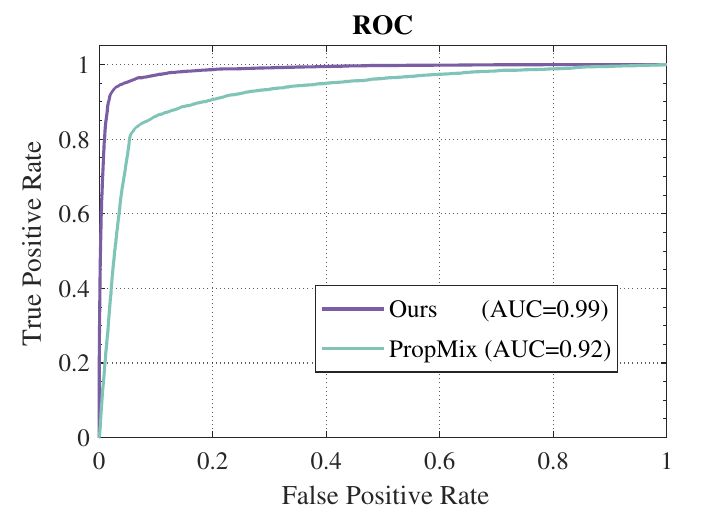}}
 \caption{AUC curves (Top) and ROC curves at 300 epoch (Bottom) on CIFAR-100 with 90\% symmetry noise.}
\label{fig:fig3}
\end{figure}

To reduce the error accumulation in sample selection, we further introduce CLIP to assist in the clean sample selection and present the collaborative sample selection with CLIP, whose results are listed in Table~\ref{tb:ablation1}(third row).
One can note that, in cooperation with CLIP, the classification performance has been promoted under all noise rates, especially under the high level of label noise, \eg, $80\%$ and $90\%$, the results improve significantly with the gain of $\sim 4\%$ and $\sim 7\%$ on CIFAR-100 in test accuracy.
This promotion is reasonable. 
For deeper analysis, we also provide the zero-shot classification performance of CLIP ($66.09\%$) on CIFAR-100. 
In contrast, the test accuracy of the baseline on CIFAR-100 is only $58.08\%$ when the noise rate is $90\%$.
High label noise levels can lead to severe overfitting on noisy samples, impacting the model's performance, and the alternating training manner makes it fall into the trap of wrong memory, thereby it could no longer distinguish the sample to clean or noisy.
CLIP is trained by using large-scale data, and has shown great potential in diverse downstream tasks.
Compared with the performance of the baseline under 90\% label noise rate on CIFAR-100, CLIP also shows its advantages via zero-shot classification.
Therefore, we are inspired to leverage CLIP for collaboration in sample selection, to relieve the confirmation bias induced by the error accumulation in sample selection.
And the results demonstrate our contribution of introducing CLIP for collaborative sample selection as well.
Furthermore, we extend the collaborative sample selection with prompt fine-tuning of CLIP and present the results in Table~\ref{tb:ablation1}(bottom row).
In the semi-supervised learning stage,  we employ a co-training mechanism to fine-tune the prompt of CLIP concurrently.
It shows that the test accuracy increases in varying degrees across different noise ratios with prompt fine-tuning.
In particular, the accuracy gap between the fixed prompt and fine-tuning prompt is $>2\%$ even in the extreme noise setting.
To prove the quality of the selected clean set, we plot the curve of the Area Under a Curve (AUC) score during training and provide the Receiver Operating Character (ROC) curve of the 300 epoch model according to the results of clean sample selection in PropMix and our collaborative sample selection method, respectively, on CIFAR-100 with the 90\%  symmetric noise in Figure~\ref{fig:fig3}.
It can be obviously seen that our method demonstrates the ability to effectively distinguish between clean and noisy labeled samples.

Moreover, we also evaluate the discriminant ability of CLIP for asymmetric noise. 
In Table~\ref{tb:similar images}, the predictions of an image of a hamster from CIFAR-100 which looks like a rabbit are listed using CLIP as an example.
It can be observed that the CLIP model accurately distinguishes the hamster with a probability of $0.448$.
Therefore,  it is evident from a visualization point that the CLIP model has great potential to generate auxiliary information for assisting clean sample selection even under asymmetric label noise.
\begin{table}
\centering
\footnotesize
\caption{The images with ground truth labels and the predicted probabilities for different labels by CLIP on similar images. }
\label{tb:similar images}
    \begin{tabular}{c|c|c|c}
    \hlinew{1.3pt}
    \makebox[0.12\textwidth][c]{GT $\backslash$ Prediction} 
    &\makebox[0.09\textwidth][c]{hamster}
    &\makebox[0.09\textwidth][c]{rabbit}
    &\makebox[0.09\textwidth][c]{others}\\
    \hline
   \makebox[0.18\textwidth][c]{\makecell[c]{ 
   {\vspace{0.5em} \hspace{33em}
    \begin{minipage}[m]{\columnwidth}
    {\centering
    \raisebox{0.2\height}{\includegraphics[width=0.7in]{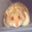}}}
    \end{minipage}
    \vspace{-1.25em}} \\
    \textcolor{blue}{\textbf{hamster}}}}
    &\makebox[0.18\textwidth][c]{0.448}
    &\makebox[0.18\textwidth][c]{0.289}
    &\makebox[0.18\textwidth][c]{0.263}\\
    \hline
    \makebox[0.18\textwidth][c]{\makecell[c]{{ 
    \hspace{33em}
    \vspace{0.3em}
    \begin{minipage}[m]{\columnwidth}
    {\raisebox{0.2\height}{\includegraphics[width=0.7in]{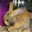}}}
    \end{minipage}
    \vspace{-1.25em}}\\
    \textcolor{blue}{\textbf{rabbit}}}}
    &\makebox[0.18\textwidth][c]{0.005}
    &\makebox[0.18\textwidth][c]{0.936} 
    &\makebox[0.18\textwidth][c]{0.059}\\
    \hlinew{1.3pt}
    \end{tabular}
\end{table}



\begin{figure}[htbp!]
\hspace{-1em}
\subfloat{
\includegraphics[width=0.45\linewidth]{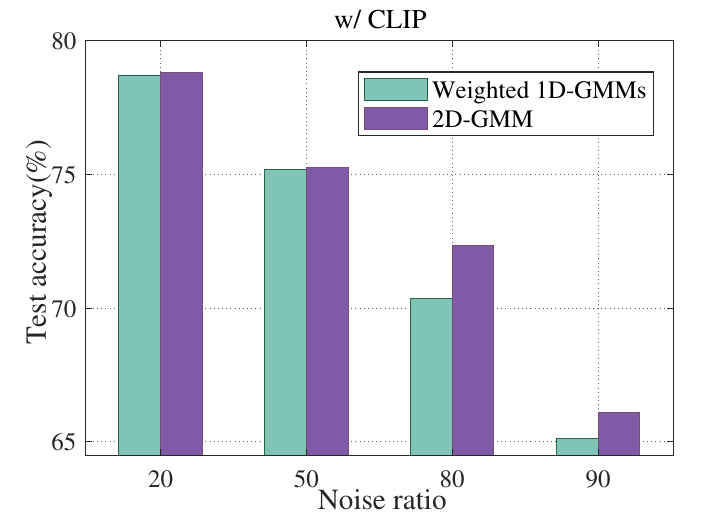}}
\quad
\subfloat{
\includegraphics[width=0.45\linewidth]{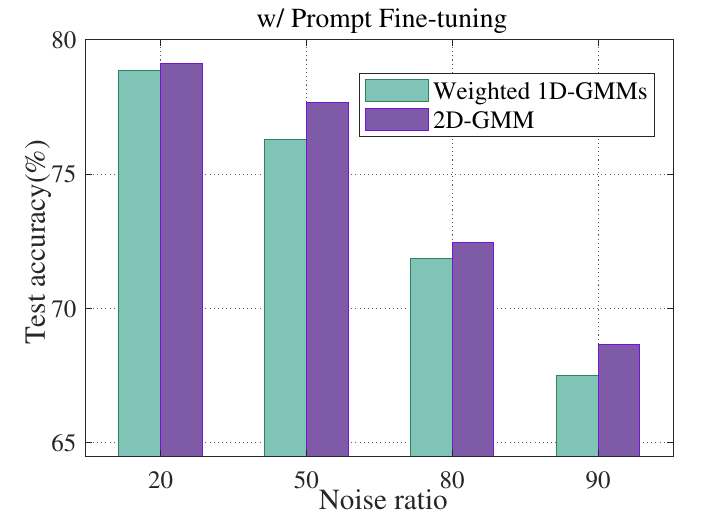}}
 \caption{Ablation study in terms of test accuracy (\%) on CIFAR-100 to evaluate the effect of the Collaborative Sample Selection (CSS) scheme. CSS with CLIP (Top) and CSS with prompt fine-tuning (Bottom).}
\label{fig:fig4}
\end{figure}

%
\noindent \textbf{Effect of Collaborative Sample Selection Scheme.} 
To aggregate the probabilities from the classifier and CLIP, we train a two-component 2D-GMM to implement collaborative sample selection.
Traditional approaches usually fit the two-component 1D-GMM to cross-entropy values and obtain the clean probability for each sample.
To demonstrate the superiority of our 2D-GMM model, we compare it with the most intuitive and simplest fusion manner, \ie, the weighted clean probability from two 1D-GMMs (weighted 1D-GMMs).
For $\omega_{i,1}$ and $\omega_{i,2}$ from DNNs and CLIP respectively, we use the fixed coefficient to compute the weighted clean probability for sample $\mathbf{x}_i$, \ie,  $ p(\boldsymbol{\omega}_i) = \beta \cdot p_{\text{\tiny GMM}}(\omega_{i,1}) +(1-\beta)\cdot p_{\text{\tiny GMM}}(\omega_{i,2})$, where $p_{\text{\tiny GMM}}(\cdot)$ denotes the clean probability obtained from the two-component 1D-GMM model, and $\beta$ is the weight coefficient set as $0.2$.
When the prompt is fixed, all training samples are determined by the predictions of CLIP.
%
From Figure~\ref{fig:fig4}, our 2D-GMM method improves the performance significantly over the weighted 1D-GMMs under all noise rates.
Furthermore, we also validate the aggregating method with prompt fine-tuning.
In Figure~\ref{fig:fig4}, the test accuracy of the weighted 1D-GMMs is 67.51\% and that of 2D-GMM is 68.66\% at $90\%$ noise rate.
Results show that the dynamic method (2D-GMM) exhibits better performance.

\noindent \textbf{Effect of Loss Functions.} 
To study the effect of loss functions in semi-supervised learning, we remove each term to provide more insights for SSL.
First, we discard the loss of unlabeled samples $\mathcal{L}_{\mathcal{U}}^{\rm D}$ in Eq.~(\ref{eq:DNNs}), \ie, w/o $\mathcal{L}_{\mathcal{U}}^{\rm D}$ in Table~\ref{tb:term in loss}.
Without the inclusion of the unlabeled loss, the test accuracy on CIFAR-100 experiences varying degrees of decline across different noise ratios.
Especially when encountering a high level of label noise, \eg, $90\%$, the results exhibit a significant decline, with a decrease of approximately $15\%$ on CIFAR-100.
Since a large quantity of useful information from unlabeled samples is discarded, especially when the noise level is extremely high, the classification performance will inevitably degrade without $\mathcal{L}_{\mathcal{U}}^{\rm D}$ in the SSL stage.
As for the contrastive loss, we have demonstrated its effectiveness in Table~\ref{tb:term in loss} compared with the baseline. 
Here, deployed with the proposed collaborative sample selection, we remove the contrastive loss $\mathcal{L}_{con}$ in SSL, and achieve the variant, \ie, w/o $\mathcal{L}_{con}$ in Table~\ref{tb:term in loss}.
Notably, even with the improved sample selection scheme, removing $\mathcal{L}_{con}$ also results in a slight drop in classification performance on the overall symmetry noise ratios.
Moreover, we also employ the regularization term $\mathcal{L}_{reg}$ in SSL to prevent assigning all samples to a single class.
Therefore, we carry out experiments by getting rid of the regularization term, \ie, w/o $\mathcal{L}_{reg}$, to evaluate its impact.
As shown in Table~\ref{tb:term in loss}, we note that with the increase in noise ratio, the influence of the regularization term on the classifier also increases.

\noindent \textbf{Effect of Prompt Fine-tuning Mechanism.} 
Many recent works have demonstrated the superiority of prompt fine-tuning of CLIP as in ~\citep{zhou2022learning, zhou2022conditional}.
To adapt the pre-trained CLIP to the task of learning with noisy labels, we also learn the prompt jointly with the DNNs classifier in the SSL stage.
Here, to study the effect of different learning methods for fine-tuning the prompt of CLIP, we provide three variants, \ie, Zero-shot selecting, Prompt fine-tuning by SL, and Prompt fine-tuning by SSL (Ours).
In Table~\ref{tb:ablation4}, we first provide the test accuracy of Zero-shot selection on CIFAR-100, which refers to the classification result using CLIP only with a fixed prompt.
Then, we fine-tune the prompt by the  supervised loss $\mathcal{L}_{\mathcal{X}}^{\rm P}$ in Eq.~(\ref{eq:label loss of prompt}) with the selected clean set, leading to the variant, \ie, Prompt fine-tuning by SL.
It can be noted that, its test accuracy outperforms the results of zero-shot selecting with the fixed prompt ($0.06\% \sim 1.77\% $) significantly.
Moreover, to make full use of the unlabeled samples, we apply the semi-supervision learning method with the loss $\mathcal{L}^{\rm P}$ in Eq.~(\ref{eq:prompt}) for fine-tuning the prompt of CLIP, \ie, Prompt fine-tuning by SSL (Ours), which achieves the best performance as shown in Table~\ref{tb:ablation4}.
%

%
%

%
\begin{table}[htbp!]
\footnotesize
\centering
\caption{Ablation study in terms of test accuracy (\%) on CIFAR-100 to evaluate the effect of each term in loss functions.}
 \begin{tabular}{p{2cm}<{\centering}|ccccc}
    \hlinew{1.3pt}
   \multirow{2}{*}{$\qquad$} &
   \multicolumn{4}{c}{CIFAR-100}\\
    \multicolumn{1}{c|}{$\qquad$}
    & \makebox[0.1\textwidth][c]{20\%} 
    & \makebox[0.1\textwidth][c]{50\%}
    & \makebox[0.1\textwidth][c]{80\%}
    & \makebox[0.1\textwidth][c]{90\%}\\
    \hline
    w/o $\mathcal{L}_{\mathcal{U}}^{\rm D}$
    & \makebox[0.06\textwidth][c]{75.18} & \makebox[0.06\textwidth][c]{73.92} & \makebox[0.06\textwidth][c]{61.99} & \makebox[0.06\textwidth][c]{53.90}\\
    
     w/o $\mathcal{L}_{con}$ & 
    \makebox[0.06\textwidth][c]{78.51} & \makebox[0.06\textwidth][c]{76.45} & \makebox[0.06\textwidth][c]{71.29} & \makebox[0.06\textwidth][c]{67.59}\\  
    w/o $\mathcal{L}_{reg}$ &
    \makebox[0.06\textwidth][c]{78.98} & \makebox[0.06\textwidth][c]{76.84} & \makebox[0.06\textwidth][c]{71.34} &\makebox[0.06\textwidth][c]{67.25} \\
    \hline
    Ours &
    \makebox[0.06\textwidth][c]{79.12} & \makebox[0.06\textwidth][c]{77.65} & \makebox[0.06\textwidth][c]{72.46} &\makebox[0.06\textwidth][c]{68.66} \\
    \hlinew{1.pt}
\end{tabular}\label{tb:term in loss}
\end{table}
\begin{table}[htbp!]
\footnotesize
\centering
\caption{Ablation study in terms of test accuracy (\%) on CIFAR-100 to evaluate the effect of prompt fine-tuning mechanism.}
 \begin{tabular}{l|ccccc}
    \hlinew{1.3pt}
   \multirow{2}{*}{$\qquad$} &
   \multicolumn{4}{c}{CIFAR-100}\\
    \multicolumn{1}{c|}{$\qquad$}
    & \makebox[0.1\textwidth][c]{20\%} 
    & \makebox[0.1\textwidth][c]{50\%}
    & \makebox[0.1\textwidth][c]{80\%}
    & \makebox[0.1\textwidth][c]{90\%}\\
    \hline
    Zero-shot selecting &
    \makebox[0.03\textwidth][c]{78.81} & \makebox[0.03\textwidth][c]{75.13} & \makebox[0.03\textwidth][c]{71.29} & \makebox[0.03\textwidth][c]{66.09}\\ 
    Prompt fine-tuning by SL &
    \makebox[0.03\textwidth][c]{78.95} & \makebox[0.03\textwidth][c]{76.83} & \makebox[0.03\textwidth][c]{72.41} & \makebox[0.03\textwidth][c]{67.86}\\ 
    \hline
    Prompt fine-tuning by SSL (Ours)&
    \makebox[0.03\textwidth][c]{79.12} & \makebox[0.03\textwidth][c]{77.65} & \makebox[0.03\textwidth][c]{72.46} &
    \makebox[0.03\textwidth][c]{68.66} \\
    \hlinew{1.pt}
\end{tabular}\label{tb:ablation4}
\end{table}
\section{Conclusion}
In this paper, we proposed a Collaborative Sample Selection method (CSS) method for learning with noisy labels (LNL). 
With the auxiliary information provided by the pre-trained model CLIP, our CSS could help to identify as many clean samples as possible and get rid of the mixed noisy ones from the selected clean subset, thereby ensuring the quality of the clean subset and minimizing the negative effects on subsequent semi-supervised learning.
Furthermore, to adapt CLIP to LNL and improve the feature representation of DNNs, we integrated contrastive loss and presented a co-training mechanism in the semi-supervised learning (SSL) stage to jointly fine-tune the prompt of CLIP and train the DNNs classifier.
Results on both synthetic and realistic label noise showed competitive performance with the state-of-the-art methods, and demonstrated the robustness of CSS to the over-fitting problem, especially under extremely high noise rates.

\section*{Acknowledgments}

This work is supported by the National key R\&D program of China under Grant No. 2021YFF0901203 and the National Natural Science Foundation of China (NSFC) under Grant No.62006064, No.U19A2073.





\bibliographystyle{elsarticle-num}

\bibliography{egbib}

\begin{thebibliography}{10}
\expandafter\ifx\csname url\endcsname\relax
  \def\url#1{\texttt{#1}}\fi
\expandafter\ifx\csname urlprefix\endcsname\relax\def\urlprefix{URL }\fi
\expandafter\ifx\csname href\endcsname\relax
  \def\href#1#2{#2} \def\path#1{#1}\fi

\bibitem{joseph2021towards}
K.~Joseph, S.~Khan, F.~S. Khan, V.~N. Balasubramanian, Towards open world object detection, in: Proceedings of the IEEE/CVF Conference on Computer Vision and Pattern Recognition, 2021, pp. 5830--5840.

\bibitem{krizhevsky2017imagenet}
A.~Krizhevsky, I.~Sutskever, G.~E. Hinton, Imagenet classification with deep convolutional neural networks, Communications of the ACM 60~(6) (2017) 84--90.

\bibitem{yang2021objects}
S.~Yang, P.~Sun, Y.~Jiang, X.~Xia, R.~Zhang, Z.~Yuan, C.~Wang, P.~Luo, M.~Xu, Objects in semantic topology, arXiv preprint arXiv:2110.02687 (2021).

\bibitem{ge2020cascaded}
S.~Ge, C.~Zhang, S.~Li, D.~Zeng, D.~Tao, Cascaded correlation refinement for robust deep tracking, IEEE transactions on neural networks and learning systems 32~(3) (2020) 1276--1288.

\bibitem{wang2021contrastive}
P.~Wang, K.~Han, X.-S. Wei, L.~Zhang, L.~Wang, Contrastive learning based hybrid networks for long-tailed image classification, in: Proceedings of the IEEE/CVF conference on computer vision and pattern recognition, 2021, pp. 943--952.

\bibitem{arpit2017closer}
D.~Arpit, S.~Jastrz{\k{e}}bski, N.~Ballas, D.~Krueger, E.~Bengio, M.~S. Kanwal, T.~Maharaj, A.~Fischer, A.~Courville, Y.~Bengio, et~al., A closer look at memorization in deep networks, in: International conference on machine learning, PMLR, 2017, pp. 233--242.

\bibitem{liu2020early}
S.~Liu, J.~Niles-Weed, N.~Razavian, C.~Fernandez-Granda, Early-learning regularization prevents memorization of noisy labels, in: Advances in Neural Information Processing Systems (NeurIPS), Vol.~33, 2020, pp. 20331--20342.

\bibitem{jiang2018mentornet}
L.~Jiang, Z.~Zhou, T.~Leung, L.-J. Li, L.~Fei-Fei, {M}entor{N}et: Learning data-driven curriculum for very deep neural networks on corrupted labels, in: ICML, Vol.~80, 2018, pp. 2304--2313.

\bibitem{shu2019meta}
J.~Shu, Q.~Xie, L.~Yi, Q.~Zhao, S.~Zhou, Z.~Xu, D.~Meng, {Meta-Weight-Net}: Learning an explicit mapping for sample weighting, in: Advances in Neural Information Processing Systems (NeurIPS), Vol.~32, 2019, pp. 1--12.

\bibitem{song2019selfie}
H.~Song, M.~Kim, J.-G. Lee, Selfie: Refurbishing unclean samples for robust deep learning, in: International Conference on Machine Learning, PMLR, 2019, pp. 5907--5915.

\bibitem{yi2019probabilistic}
K.~Yi, J.~Wu, Probabilistic end-to-end noise correction for learning with noisy labels, in: Proceedings of the IEEE/CVF Conference on Computer Vision and Pattern Recognition, 2019, pp. 7017--7025.

\bibitem{li2020dividemix}
J.~Li, R.~Socher, S.~C. Hoi, {DivideMix}: Learning with noisy labels as semi-supervised learning, in: ICLR, 2020.

\bibitem{propmix}
F.~R. Cordeiro, V.~Belagiannis, I.~Reid, G.~Carneiro, Propmix: Hard sample filtering and proportional mixup for learning with noisy labels, in: BMVC, 2021, p. 187.

\bibitem{zheng2021meta}
G.~Zheng, A.~H. Awadallah, S.~Dumais, Meta label correction for noisy label learning, in: Proceedings of the AAAI Conference on Artificial Intelligence, Vol.~35, 2021, pp. 11053--11061.

\bibitem{zhang2019metacleaner}
W.~Zhang, Y.~Wang, Y.~Qiao, Metacleaner: Learning to hallucinate clean representations for noisy-labeled visual recognition, in: Proceedings of the IEEE/CVF Conference on Computer Vision and Pattern Recognition, 2019, pp. 7373--7382.

\bibitem{chen2022compressing}
Y.~Chen, S.~X. Hu, X.~Shen, C.~Ai, J.~A. Suykens, Compressing features for learning with noisy labels, IEEE Transactions on Neural Networks and Learning Systems (2022).

\bibitem{karim2022unicon}
N.~Karim, M.~N. Rizve, N.~Rahnavard, A.~Mian, M.~Shah, Unicon: Combating label noise through uniform selection and contrastive learning, in: CVPR, 2022, pp. 9676--9686.

\bibitem{wei2022self}
Q.~Wei, H.~Sun, X.~Lu, Y.~Yin, Self-filtering: A noise-aware sample selection for label noise with confidence penalization, in: European Conference on Computer Vision, Springer, 2022, pp. 516--532.

\bibitem{radford2021learning}
A.~Radford, J.~W. Kim, C.~Hallacy, A.~Ramesh, G.~Goh, S.~Agarwal, G.~Sastry, A.~Askell, P.~Mishkin, J.~Clark, et~al., Learning transferable visual models from natural language supervision, in: International Conference on Machine Learning, PMLR, 2021, pp. 8748--8763.

\bibitem{li2019visualbert}
L.~H. Li, M.~Yatskar, D.~Yin, C.-J. Hsieh, K.-W. Chang, Visualbert: A simple and performant baseline for vision and language, arXiv preprint arXiv:1908.03557 (2019).

\bibitem{jia2021scaling}
C.~Jia, Y.~Yang, Y.~Xia, Y.-T. Chen, Z.~Parekh, H.~Pham, Q.~Le, Y.-H. Sung, Z.~Li, T.~Duerig, Scaling up visual and vision-language representation learning with noisy text supervision, in: International Conference on Machine Learning, PMLR, 2021, pp. 4904--4916.

\bibitem{kim2021vilt}
W.~Kim, B.~Son, I.~Kim, Vilt: Vision-and-language transformer without convolution or region supervision, in: International Conference on Machine Learning, PMLR, 2021, pp. 5583--5594.

\bibitem{patashnik2021styleclip}
O.~Patashnik, Z.~Wu, E.~Shechtman, D.~Cohen-Or, D.~Lischinski, Styleclip: Text-driven manipulation of stylegan imagery, in: Proceedings of the IEEE/CVF International Conference on Computer Vision, 2021, pp. 2085--2094.

\bibitem{li2021align}
J.~Li, R.~Selvaraju, A.~Gotmare, S.~Joty, C.~Xiong, S.~C.~H. Hoi, Align before fuse: Vision and language representation learning with momentum distillation, Advances in neural information processing systems 34 (2021) 9694--9705.

\bibitem{rao2022denseclip}
Y.~Rao, W.~Zhao, G.~Chen, Y.~Tang, Z.~Zhu, G.~Huang, J.~Zhou, J.~Lu, Denseclip: Language-guided dense prediction with context-aware prompting, in: Proceedings of the IEEE/CVF Conference on Computer Vision and Pattern Recognition, 2022, pp. 18082--18091.

\bibitem{ding2022open}
Z.~Ding, J.~Wang, Z.~Tu, Open-vocabulary panoptic segmentation with maskclip, arXiv preprint arXiv:2208.08984 (2022).

\bibitem{he2020momentum}
K.~He, H.~Fan, Y.~Wu, S.~Xie, R.~Girshick, Momentum contrast for unsupervised visual representation learning, in: Proceedings of the IEEE/CVF conference on computer vision and pattern recognition, 2020, pp. 9729--9738.

\bibitem{chen2020improved}
X.~Chen, H.~Fan, R.~Girshick, K.~He, Improved baselines with momentum contrastive learning, arXiv preprint arXiv:2003.04297 (2020).

\bibitem{chen2020simple}
T.~Chen, S.~Kornblith, M.~Norouzi, G.~Hinton, A simple framework for contrastive learning of visual representations, in: International conference on machine learning, PMLR, 2020, pp. 1597--1607.

\bibitem{chen2020big}
T.~Chen, S.~Kornblith, K.~Swersky, M.~Norouzi, G.~E. Hinton, Big self-supervised models are strong semi-supervised learners, Advances in neural information processing systems 33 (2020) 22243--22255.

\bibitem{song2022learning}
H.~Song, M.~Kim, D.~Park, Y.~Shin, J.-G. Lee, Learning from noisy labels with deep neural networks: A survey, IEEE Transactions on Neural Networks and Learning Systems (2022).

\bibitem{sukhbaatar2014training}
S.~Sukhbaatar, J.~Bruna, M.~Paluri, L.~Bourdev, R.~Fergus, Training convolutional networks with noisy labels, arXiv preprint arXiv:1406.2080 (2014).

\bibitem{chang2017active}
H.-S. Chang, E.~Learned-Miller, A.~McCallum, Active bias: Training more accurate neural networks by emphasizing high variance samples, Advances in Neural Information Processing Systems 30 (2017).

\bibitem{goldberger2016training}
J.~Goldberger, E.~Ben-Reuven, Training deep neural-networks using a noise adaptation layer, in: International conference on learning representations, 2016.

\bibitem{lyu2019curriculum}
Y.~Lyu, I.~W. Tsang, {Curriculum Loss}: Robust learning and generalization against label corruption, in: International Conference on Learning Representations (ICLR), 2020.

\bibitem{ghosh2017robust}
A.~Ghosh, H.~Kumar, P.~Sastry, Robust loss functions under label noise for deep neural networks, in: Proceedings of the Thirty-First {AAAI} Conference on Artificial Intelligence (AAAI), Vol.~31, 2017, pp. 1919--1925.

\bibitem{zhang2018generalized}
Z.~Zhang, M.~Sabuncu, Generalized cross entropy loss for training deep neural networks with noisy labels, Advances in neural information processing systems 31 (2018).

\bibitem{han2018co}
B.~Han, Q.~Yao, X.~Yu, G.~Niu, M.~Xu, W.~Hu, I.~Tsang, M.~Sugiyama, Co-teaching: Robust training of deep neural networks with extremely noisy labels, in: NIPS, Vol.~31, 2018, pp. 8535--8545.

\bibitem{sharma2020noiserank}
K.~Sharma, P.~Donmez, E.~Luo, Y.~Liu, I.~Z. Yalniz, Noiserank: Unsupervised label noise reduction with dependence models, in: ECCV, Springer, 2020, pp. 737--753.

\bibitem{li2022selective}
S.~Li, X.~Xia, S.~Ge, T.~Liu, Selective-supervised contrastive learning with noisy labels, in: CVPR, 2022, pp. 316--325.

\bibitem{ortego2021multi}
D.~Ortego, E.~Arazo, P.~Albert, N.~E. O'Connor, K.~McGuinness, Multi-objective interpolation training for robustness to label noise, in: CVPR, 2021, pp. 6606--6615.

\bibitem{jiang2022sparse}
R.~Jiang, Y.~Yan, J.-H. Xue, B.~Wang, H.~Wang, When sparse neural network meets label noise learning: A multistage learning framework, IEEE Transactions on Neural Networks and Learning Systems (2022).

\bibitem{ma2021learning}
F.~Ma, Y.~Wu, X.~Yu, Y.~Yang, Learning with noisy labels via self-reweighting from class centroids, IEEE Transactions on Neural Networks and Learning Systems 33~(11) (2021) 6275--6285.

\bibitem{yao2021jo}
Y.~Yao, Z.~Sun, C.~Zhang, F.~Shen, Q.~Wu, J.~Zhang, Z.~Tang, Jo-src: A contrastive approach for combating noisy labels, in: Proceedings of the IEEE/CVF Conference on Computer Vision and Pattern Recognition, 2021, pp. 5192--5201.

\bibitem{zhou2022learning}
K.~Zhou, J.~Yang, C.~C. Loy, Z.~Liu, Learning to prompt for vision-language models, International Journal of Computer Vision 130~(9) (2022) 2337--2348.

\bibitem{zhou2022conditional}
K.~Zhou, J.~Yang, C.~C. Loy, Z.~Liu, Conditional prompt learning for vision-language models, in: Proceedings of the IEEE/CVF Conference on Computer Vision and Pattern Recognition, 2022, pp. 16816--16825.

\bibitem{grill2020bootstrap}
J.-B. Grill, F.~Strub, F.~Altch{\'e}, C.~Tallec, P.~Richemond, E.~Buchatskaya, C.~Doersch, B.~Avila~Pires, Z.~Guo, M.~Gheshlaghi~Azar, et~al., Bootstrap your own latent-a new approach to self-supervised learning, Advances in neural information processing systems 33 (2020) 21271--21284.

\bibitem{permuter2006study}
H.~Permuter, J.~Francos, I.~Jermyn, A study of gaussian mixture models of color and texture features for image classification and segmentation, Pattern recognition 39~(4) (2006) 695--706.

\bibitem{d2021consistency}
A.~D'Ortenzio, C.~Manes, Consistency issues in gaussian mixture models reduction algorithms, arXiv preprint arXiv:2104.12586 (2021).

\bibitem{moon1996expectation}
T.~K. Moon, The expectation-maximization algorithm, IEEE Signal processing magazine 13~(6) (1996) 47--60.

\bibitem{sohn2020fixmatch}
K.~Sohn, D.~Berthelot, N.~Carlini, Z.~Zhang, H.~Zhang, C.~A. Raffel, E.~D. Cubuk, A.~Kurakin, C.-L. Li, {FixMatch}: Simplifying semi-supervised learning with consistency and confidence, in: Advances in Neural Information Processing Systems (NeurIPS), Vol.~33, 2020, pp. 596--608.

\bibitem{khosla2020supervised}
P.~Khosla, P.~Teterwak, C.~Wang, A.~Sarna, Y.~Tian, P.~Isola, A.~Maschinot, C.~Liu, D.~Krishnan, Supervised contrastive learning, Advances in Neural Information Processing Systems 33 (2020) 18661--18673.

\bibitem{krizhevsky2009learning}
A.~Krizhevsky, G.~Hinton, et~al., Learning multiple layers of features from tiny images (2009).

\bibitem{xiao2015learning}
T.~Xiao, T.~Xia, Y.~Yang, C.~Huang, X.~Wang, Learning from massive noisy labeled data for image classification, in: Proceedings of the IEEE conference on computer vision and pattern recognition, 2015, pp. 2691--2699.

\bibitem{li2017webvision}
W.~Li, L.~Wang, W.~Li, E.~Agustsson, L.~Van~Gool, Webvision database: Visual learning and understanding from web data, arXiv preprint arXiv:1708.02862 (2017).

\bibitem{deng2009imagenet}
J.~Deng, W.~Dong, R.~Socher, L.-J. Li, K.~Li, L.~Fei-Fei, Imagenet: A large-scale hierarchical image database, in: 2009 IEEE conference on computer vision and pattern recognition, Ieee, 2009, pp. 248--255.

\bibitem{arazo2019unsupervised}
E.~Arazo, D.~Ortego, P.~Albert, N.~O’Connor, K.~McGuinness, Unsupervised label noise modeling and loss correction, in: International conference on machine learning, PMLR, 2019, pp. 312--321.

\bibitem{jiang2020beyond}
L.~Jiang, D.~Huang, M.~Liu, W.~Yang, Beyond synthetic noise: Deep learning on controlled noisy labels, in: International Conference on Machine Learning, PMLR, 2020, pp. 4804--4815.

\bibitem{patrini2017making}
G.~Patrini, A.~Rozza, A.~Krishna~Menon, R.~Nock, L.~Qu, Making deep neural networks robust to label noise: A loss correction approach, in: Proceedings of the IEEE conference on computer vision and pattern recognition, 2017, pp. 1944--1952.

\bibitem{iscen2022learning}
A.~Iscen, J.~Valmadre, A.~Arnab, C.~Schmid, Learning with neighbor consistency for noisy labels, in: Proceedings of the IEEE/CVF Conference on Computer Vision and Pattern Recognition, 2022, pp. 4672--4681.

\bibitem{he2016identity}
K.~He, X.~Zhang, S.~Ren, J.~Sun, Identity mappings in deep residual networks, in: European conference on computer vision, Springer, 2016, pp. 630--645.

\bibitem{malach2017decoupling}
E.~Malach, S.~Shalev-Shwartz, Decoupling" when to update" from" how to update", Advances in neural information processing systems 30 (2017).

\bibitem{zhang2020symmetry}
Y.~Zhang, Q.~Qu, J.~Wright, From symmetry to geometry: Tractable nonconvex problems, arXiv preprint arXiv:2007.06753 (2020).

\bibitem{yi2022learning}
L.~Yi, S.~Liu, Q.~She, A.~I. McLeod, B.~Wang, On learning contrastive representations for learning with noisy labels, in: Proceedings of the IEEE/CVF Conference on Computer Vision and Pattern Recognition, 2022, pp. 16682--16691.

\end{thebibliography}

\end{document}